\documentclass[11pt]{article}

\usepackage{amsmath, xparse}
\usepackage{bm}             

\usepackage[nonatbib,preprint]{neurips_2025}

\usepackage[utf8]{inputenc} 
\usepackage[T1]{fontenc}    
\usepackage{hyperref}       
\usepackage{url}            
\usepackage{booktabs}       
\usepackage{amsfonts}       
\usepackage{nicefrac}       
\usepackage{microtype}      
\usepackage{xcolor}         
\usepackage{graphicx}

\usepackage[numbers]{natbib}

\title{Reinforcement Learning as (Discrete) Potential Theory}
\author{Chris Connolly\thanks{Elements of this paper were prepared with assistance from SRI's subscription to Chat-GPT 5.2.}\\
Computer Science Laboratory\\
SRI International}
\date{April 28, 2026}

\begin{document}

\maketitle

\bibliographystyle{plain}

\begin{abstract}

  Reinforcement learning (RL) theory fundamentally depends on
  probability theory through the Markov chain.  There is a deep
  connection between probability theory and potential theory.  This
  paper reviews that connection and explores the potential-theoretic
  viewpoint for core reinforcement learning representations and
  algorithms under a fixed-policy assumption.  This viewpoint may
  offer a path for improved sample efficiency and formal constraints
  that can be applied to RL.  When the fixed-policy assumption is
  relaxed, the linear potential theory framework can be naturally
  extended to the nonlinear case.\footnote{Although the material here
  is presented in terms of {\em discrete} potential theory, the
  extension to continuous state spaces follows from the connection to
  potential theory.}

\end{abstract}

\section{Introduction}

Reinforcement learning (RL) theory fundamentally depends on
probability theory through the Markov chain.  There is a deep
connection between probability theory and potential theory.  This
paper is primarily a review that examines RL through the lens of
potential theory.  While many of the concepts described here represent
decades of research, the equivalences between probability theory and
potential theory offer viewpoints that may help improve the state of
the art in reinforcement learning, for example sample efficiency and
credit assignment.  Potential theory and probability theory (and
therefore RL) are fundamentally linked in the sense that Markov chain
outcome probabilities are equivalent to potentials that are
encountered in natural physical phenomena \cite{Doob,Kemeny76}.  This
equivalence was explored by Doob and is perhaps most clearly and
simply described by Doyle and Snell in terms of resistive networks
\cite{Doyle84}: a Markov chain state corresponds to a node in an
equivalent resistive network, and each resistor in that network
represents an entry in the Markov chain transition matrix.  With this
equivalence in mind, there are three cases of fixed-policy\footnote{We
fix policy so that we can maintain linearity and more clearly
demonstrate the direct equivalences.  This assumption can be relaxed
by expressing the nonlinear (e.g., policy optimization) problem in
terms of a decomposition into multiple linear subproblems.}
Reinforcement Learning (RL) that can be expressed in terms of
potential theory:

\begin{enumerate}
\item \label{item:first} Episodic RL\footnote{``Episodic'' is synonymous with ``absorbing''.} with reward only at terminal states $\Longleftrightarrow$ Laplace's Equation
\item \label{item:second} Ongoing control with distributed reward $\Longleftrightarrow$ Poisson's Equation
\item \label{item:third} RL problems with nonstationary environments $\Longleftrightarrow$ The Diffusion (Heat) Equation
\end{enumerate}
All three cases admit both discrete and continuous domains.  Finite
difference methods, for example, can be seen as discrete potential
problems.  The potential-theoretic view of RL allows us to consider
optimal state-action sequences as {\em streamlines} in a flow induced
by the value function: the flow of value through state space.
For clarity's sake most of this discussion will
focus on discrete problems on graphs, intuitions for which can be
found in Doyle and Snell \cite{Doyle84}.

Case \ref{item:first} is associated with boundary value problems for
Laplace's Equation (and Markov chains).  These problems are defined
solely with respect to boundary conditions (absorbing states).  Case
\ref{item:second} is associated with non-absorbing (ongoing) control
problems, usually unbounded in time with {\em distributed} reward
(e.g. the cart-pole balancing problem).  Case \ref{item:third} is associated
with Markov decision processes in {\em nonstationary} environments,
i.e. with time-varying transient and boundary conditions.  Cases
\ref{item:first} and \ref{item:second} are the equilibrium asymptotes
for case \ref{item:third}.

The connection between Markov chains and potential theory is well
known within certain communities: harmonic functions, Poisson
equations, and Green's functions\footnote{In potential theory, a
Green's Function maps boundary conditions to interior solutions.}
correspond to hitting probabilities, expected costs, and occupancy
measures of Markov chains respectively \cite{Doyle84, LevinPeres2017}.
In RL, these structures appear directly in value functions and
temporal-difference learning \cite{Barto98,Puterman94}.  The
potential-theoretic view has value in identifying formal properties of
RL structures like value or Q functions, especially in the absorbing
case.  Furthermore, the equivalence suggests that physical (analog)
substrates offer alternatives to digital solutions.  Features of RL
appear to have homologues in the neurobiology of action selection; the
most pronounced being the dopamine Reward Prediction Error (RPE)
signal discovered and researched by Wolfram Schultz \cite{SchultzDM97}
among others.  With this homologue in mind, it is tempting to
speculate a relationship between Dayan's Successor
Representation\footnote{The Successor Representation is essentially a
Green's Function for a Markov chain.  It allows one to compute the
value function by evaluating the product of the SR matrix and the
reward function.}\cite{Dayan93} and the presence of electrotonic
coupling\footnote{A neuroscientific term that includes ``resistive
coupling'', but more generally a coupling that also admits ions and
small molecules.} in the striatum, possibly offering a fast diffusion
mechanism for mapping a reward landscape into a value function.
Astrocyte\footnote{Astrocytes are glial cells that can mediate ionic
stabilization of the extracellular milieu.  Recent research suggests
that they may serve a critical computational role in cognition.}
networks could provide the required diffusion substrate.

\subsection{Preliminaries}

\subsubsection{Laplace, Poisson, and Heat Equations}

In potential theory, a {\em potential function} $v(x)$ (e.g., temperature,
electrical potential, fluid potential) is a function over a variable
in some domain $\mathcal{S}$ that is typically a subset of the reals:
$v(x), x \in S \subset R^n$.  The function $v(x)$ satisfies a partial
differential equation (PDE) subject to boundary conditions, source
conditions, or initial conditions (or combinations thereof).  The
Laplacian operator is written:
\begin{equation}
  \nabla^2 := \sum_i \frac{\partial^2}{\partial x_i^2}
\end{equation}
and is the spatial diffusion term.

The three equations we discuss here are Laplace's Equation:
\begin{equation}
  \nabla^2 v(x) = 0
\end{equation}

Poisson's Equation:
\begin{equation}
  \label{eqn:poisson}
  \nabla^2 v(x) = f(x)
\end{equation}

and the Heat Equation, sometimes known as the Diffusion Equation:
\begin{equation}
\nabla^2 v(x,t) = \alpha \frac{\partial v(x,t)}{\partial t}
\end{equation}

Laplace's equation models physical phenomena when there are no
internal sources in the interior of the domain $\mathcal{S}$.
Constraints are imposed only at the boundary $\partial \mathcal{S}$ of
the domain.  Laplace's equation is used to model steady-state
potentials (electical, fluid, thermal, etc.), and yields
minimum-energy, maximum-entropy functions.  Solutions to Laplace's
equation exist and are unique when proper boundary conditions are
established; solutions are also known as {\em harmonic functions}.
Harmonic functions exhibit the {\em mean-value} property.  This means
that the value of the function at any point is the (possibly weighted)
{\em mean} of the values at neighboring points.  One might, for
example, establish temperature boundary conditions on the edges of a
hot plate.  The interior of the plate will then reach an equilibrium
such that the temperature on the plate as a function of position is a
harmonic function.  The same is true of resistive networks and
irrotational, frictionless fluid potentials.

Poisson's equation admits a source function $f(x)$ that results in
solutions that are not harmonic, but correspond well to RL problems
that involve distributed reward.  Poisson's equation extends Laplace's
equation by considering physical phenomena that involve ongoing (but
stationary) perturbation of the diffusion equilibrium, such as current,
fluid or heat sources.  Solutions to Poisson's equation still
typically represent steady-state phenomena.  Solutions to
Poisson's equation exist and are unique up to an additive constant.
One can enforce uniqueness by constraining solutions to be zero-mean.

Finally, the Heat Equation allows us to consider time-varying
conditions subject to a diffusivity constant $\alpha$.  When the
time-derivative vanishes, we are left with one of Laplace's Equation,
Poisson's Equation, or a combination (superposition) of the two.
representing the equilibrium solution.

\subsubsection{The MDP}

A discrete-time Markov Decision Process (MDP) is a core structure in
RL, and is a mathematical model for sequential decision-making where
the next state depends only on the current state and chosen action
(the Markov property).  The MDP can be defined as a tuple:
$(\mathcal{S}, \mathcal{A}, P, r, \gamma)$ where:
\begin{itemize}

\item $\mathcal{S}$ is a set of states.

\item $\mathcal{A}$ is a set of actions.

\item $P$ is a transition operator $P: \mathcal{S} \times \mathcal{A}
  \rightarrow \mathcal{S}$ describes the transition probability
  $P(s'|s,a)$ that a process in state $s$ executing action $a$ will
  transition to state $s'$.

\item $r$ is a per-state reward.

\item $\gamma$ is a discounting factor for reward expectation.

\end{itemize}

We partition $\mathcal{S}$ into an absorbing set $\partial
\mathcal{S}$ and the set of transient states $\mathcal{T} =
\mathcal{S} \setminus \partial \mathcal{S}$.  Generally, absorbing
sets are useful for episodic RL, while $\partial \mathcal{S} =
\emptyset$ is possible for ongoing RL control problems.  Hybrid MDPs
are also possible.  For clarity we assume a fixed policy $\pi$ to
establish the basic equivalences, initially without explicit reference
to action spaces.  A policy $\pi(a|s)$ is the probability of selecting
an action $a$ given a state $s$.  A policy can be deterministic.
Since $P$ is a stochastic matrix over an augmented state-action space,
we can always marginalize over actions at each state to construct a
stochastic transition operator\footnote{Recall that each row of a {\em stochastic
  matrix} sums to 1.} for the pure state space.

Within $\partial\mathcal{S}$, we can consider two subsets\footnote{We
choose two subsets for simplicity's sake, but a much greater variety
of boundary conditions can be chosen, based on problem constraints and
desired outcomes.}  $\mathcal{A}\subset\partial\mathcal{S}$ and
$\mathcal{B}\subset\partial\mathcal{S}$, $\mathcal{A}\cup\mathcal{B} =
\partial\mathcal{S}$ and the value function $V(s)$ over states $s \in
\mathcal{S}$ in the MDP.  We assign $V(s)=1, s \in \mathcal{A}$ and
$V(s)=0, s \in \mathcal{B}$.  These provide the {\em boundary
  conditions} for the value function $V(s)$ over all states.  The
value function resulting from this set of conditions is exactly {\em
  the probability that a Markov process starting at any state s
  reaches set} $\mathcal{A}$ {\em before reaching} $\mathcal{B}$.

Transition matrices for both episodic and ongoing RL admit a form of
Dayan's Successor Representation, a linear operator $\bf{M}$ that maps
reward vectors directly to value functions $V$
\cite{Dayan93}. Gradient ascent on $V$ maximizes success
probability.  When thinking of this problem in terms of success
probability, we assume that the reward function $r$ is normalized to
the interval $[0,1]$, and the resulting value function provides the
probabiity of success at each state.  If this normalization
constraint is relaxed, the value function is an expected {\em value}
that might represent a payoff, for example.

A fixed-policy Markov Decision Process (MDP) is a Markov Reward Process (MRP) with
transition matrix $P = P_\pi$ and expected one-step reward vector $r = r_\pi$.
The value function is the expectation
\begin{equation}
V^\pi(s) = \mathbb{E}_\pi \left[ \sum_{t\ge0} \gamma^t R_{t+1} \mid S_0=s \right]
\end{equation}
and satisfies the Bellman equation\footnote{Our treatment uses $V$ for RL value and $v$ when the
context is potential theory: $V \equiv v$ is the potential and value function over
state.}
\begin{equation}
  \label{eqn:tdgamma}
V = r + \gamma P V.
\end{equation}
where the discount factor $\gamma \in [0,1]$.  Equivalently,
\begin{equation}
(I - \gamma P)V = r.
\label{eqn:disc-poisson}
\end{equation}

The linear system in Equation \ref{eqn:disc-poisson} is the discrete
analogue of Poisson's equation (Equation \ref{eqn:poisson}) where the reward function $r(x) = f(x)$
plays the role of the source function $f(x)$ in Poisson's Equation.
Equation \ref{eqn:tdgamma}, while it is assumed to be in matrix form
here, is also the formula for a single step of Barto and Sutton's {\em
  Temporal Differencing} (TD) credit assignment algorithm \cite{Barto98}.

\begin{figure}[h]
  \centering
  \includegraphics[width=0.8\textwidth]{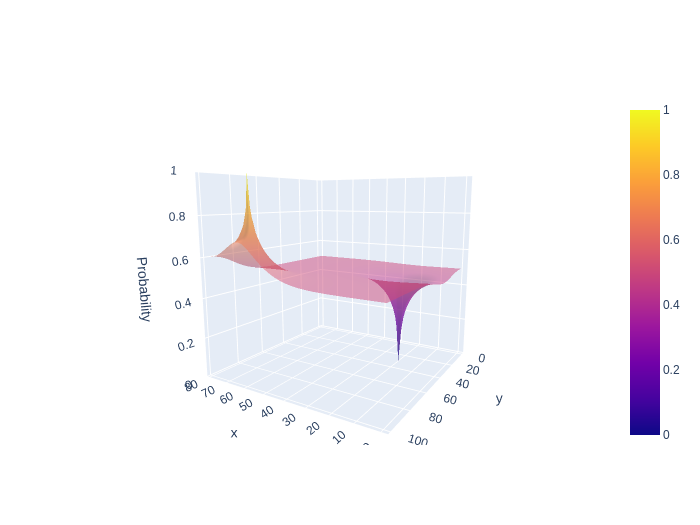}
  \caption{Discrete harmonic value function on an 80x80 grid domain where absorbing states are shown with P=1 and P=0.
    In this example, values at grid edges are allowed to float.}
  \label{fig:harmonic}
\end{figure}  

\section{Case by Case Equivalences}

\subsection{Discrete Laplace Equation: Harmonic Value Functions}

Harmonic functions are solutions to Laplace's equation {\em and} are
right eigenvectors for absorbing Markov chains.  Laplace's equation is
a model for an equilibrium potential over a domain whose interior has
no sources.  For example, Laplace's equation describes ideal fluid
flow in a domain: low-probability states are sources, and
high-probability states are sinks.  A flow defined by this potential
yields {\em streamlines}.  Following a streamline maximizes the probability of reaching a sink.  Laplace's equation
constrains to zero the sum of second derivatives of the value function $v(x)$:
\begin{equation}
  \label{lap2}
  \nabla^2 v = 0
\end{equation}
subject to boundary conditions.  When $x$ is a discrete lattice,
Equation \ref{lap2} can be turned into a finite difference equation,
where $v(x_k)$ for some $k$ is simply the weighted average of
neighboring points on the lattice.  If we use a transition matrix $P$ to represent
the averaging kernel, we can define the {\em discrete} Laplacian as:
\begin{equation}
\nabla^2 := I - P.
\end{equation}

A function $v$ is harmonic if
\begin{equation}
  \label{recur}
v = P v 
\quad \Longleftrightarrow \quad 
\nabla^2 v = 0.
\end{equation}

Fixed-point iteration of $P$ on $v$ is equivalent to a Jacobi or
Gauss-Seidel iteration for numerically solving the corresponding
potential problem.  The action of $P$ replaces each state's value with
the weighted average of neighboring states' values.  This process
converges by virtue of the fact that $P$ is stochastic.

Note that $v$ is the first {\em right} eigenvector of the transition
operator $P$.  In contrast to Equation \ref{recur}, the
{\em left} eigenvector:
\begin{equation}
\pi = \pi P
\end{equation}
corresponds to the {\em stationary distribution} of the transition
operator.  The stationary distribution provides the {\em visit rates}
of a non-absorbing Markov chain.  Generally, the Poisson case
(distributed reward) admits a nontrivial stationary distribution but a
trivial hitting probability (unless some states are absorbing).  The
Laplace case (absorbing) admits trivial stationary distributions:
visit rates are concentrated at absorbing states (with transient
states' distributions vanishingly small), but the hitting
probabilities are nontrivial or constant.

\subsection*{RL Interpretation (Terminal-Only Reward)}

Let $\mathcal{S}$ be the set of all states of an absorbing Markov chain and let $\partial \mathcal{S}$ be the set of absorbing states.
Let $\mathcal{T} = \mathcal{S} \setminus \partial \mathcal{S}$
be the set of {\em transient} (interior) states.\footnote{That is, the set difference of S and its absorbing states.}
In an episodic MDP where reward occurs only at terminal states $\partial \mathcal{S}$, {\em transient} states
 $s \in \mathcal{T}$ satisfy
\begin{equation}
V(s) = \sum_{s'\in \mathcal{S}} P_\pi(s,s') V(s').
\end{equation}
or in vector form:
\begin{equation}
  v = P v
\end{equation}
Thus $V$ (or $v$) is harmonic on transient states, with Dirichlet boundary condition
\begin{equation}
V(s) = g(s), \quad s \in \partial \mathcal{S}.
\end{equation}
where g(s) is the reward value when absorption occurs.

This is precisely the discrete Dirichlet problem, where boundary
conditions can be interpreted probabilistically.  For example, suppose
$\partial \mathcal{S}$ is partitioned into two sets, $W$ and $L$, such
that:
\begin{eqnarray}
  g(x) = 1, \quad x & \in W \\
  g(y) = 0, \quad y & \in L
\end{eqnarray}
Then the value function at any transient state provides the
probability that a random process will hit a state in set $W$ (win)
before hitting any state in set $L$ (lose).  Markov chain harmonic
functions are therefore also called ``hitting probabilities''
\cite{Kemeny76}.  A deterministic policy that seeks to maximize the
probability of success will therefore ascend the value gradient to
reach set $W$.  Effectively, this recipe follows {\em the flow of
  value} in {\em streamlines} that maximize reward.

\subsection*{Formal Properties of Harmonic Functions}

Any  value function of an absorbing MDP is harmonic.  This establishes
some strong necessary conditions for value functions:
\begin{enumerate}
  \item {\bf Mean Value Property:} The value at any transient state is
    the (weighted) mean of values at neighboring states.
  \item {\bf Min-Max Property:} In any neighborhood $\mathcal{N}$ of a
    transient state, the value function achieves its minimum and
    maximum value {\em only} on the boundary $\partial \mathcal{N}$ of that neighborhood.
\end{enumerate}

It is obvious that all linear functions are also harmonic.  If
$g(x)=c$ for all boundary states ${x \in \partial \mathcal{S}}$, then
the resulting value function must also be the constant c.  These
properties suggest strong convergence tests for value functions.  In
the absorbing case, if a TD-derived value function exhibits local
extrema in the set of transient states, then it has not converged
properly.  Likewise, the residual $Pv_n - v_{n+1}$ provides a kind of
convergence score that indicates whether more training is needed.

\subsection{Discrete Poisson Equation: Rewards as Sources}

The Poisson variant of RL admits distributed rewards that are not
necessarily limited to absorbing states.  This corresponds to a
physical problem with sources (charge, fluid, etc.) in the interior of
the domain.  In the RL case, sources are rewards.  The Poisson case
corresponds to ongoing optimal control problems of the sort that are
familiar to the RL community in environments like the cart-pole
balancing problem.

\begin{figure}[h]
  \centering
  \includegraphics[trim={1cm 0cm 0cm 0cm}, clip, width=1\textwidth]{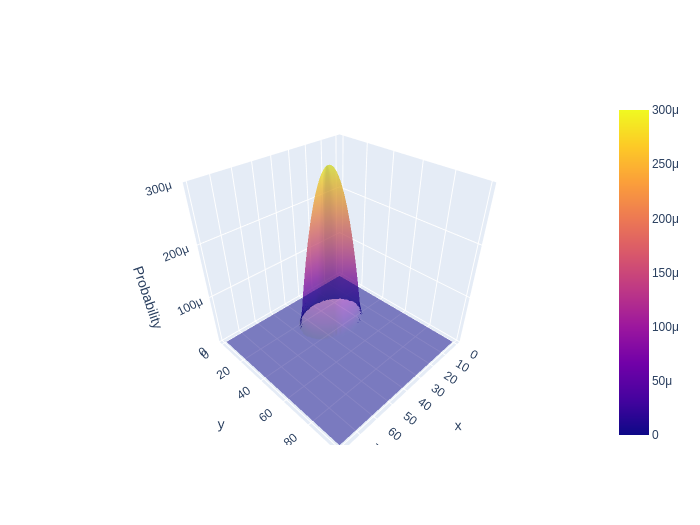}
  \caption{A distributed reward function for a Poisson-style RL problem.}
  \label{fig:poisson_reward}
\end{figure}

With per-step reward $r$, the undiscounted Bellman equation becomes
\begin{eqnarray}
  V = r + P V \\
  (I - P) V = r
\end{eqnarray}
or
\begin{equation}
\nabla^2 V = r.
\end{equation}

This is the discrete Poisson equation: the reward acts as a
distributed source term.

\subsection*{Discounted Case}

For $\gamma < 1$,
\begin{equation}
(I - \gamma P)V = r.
\label{eqn:resolvent}
\end{equation}
The introduction of distributed rewards that do not involve absorbing
sets results in a process that is unbounded in time and can be used to
solve persistent control problems (e.g., cart-pole balancing).

Equation \ref{eqn:resolvent} corresponds to a ``killed'' or discounted
Markov chain \cite{LevinPeres2017, Puterman94}.  This in turn
corresponds to a subharmonic function, but begs the interpretation of
discounting in the context of potential theory.  For the Laplace case
in three dimensions, a point source away from boundary conditions will
exhibit a potential that roughly obeys the inverse square law, so a
hyperbolic discounting is immediately available in the
potential-theoretic formulation when $\gamma = 1$.\footnote{This has
implications for modeling the exact nature of discounting of future
rewards in cognitive neuroscience experiments.}

\subsection{The Heat Equation and Temporal Differencing}

The Heat Equation represents a more general expression of the RL
problem, and furthermore admits nonstationary environments.  The
classical Heat Equation relates the first time derivative of value to
the spatial Laplacian -- it says that the difference of $v(x,t)$ in
time ($v(x,t+1)-v(x,t)$) is equal to the spatial difference between the
spatial mean of $v$ and its current value at all states:
\begin{equation}
\frac{\partial v}{\partial t} = \alpha \nabla^2v
\end{equation}
describing smoothing of value over time\footnote{The Black-Scholes options
pricing model is a Heat Equation with an advection term.}, where
$\alpha$ is equivalent to {\em thermal diffusivity} of a material.
The value function in this case can be defined subject to boundary
conditions, initial conditions, and time-varying sources.

For RL, iterative policy evaluation
\begin{equation}
V_{k+1} = r + \gamma P V_k
\label{eqn:policy}
\end{equation}

is a relaxation process that in the time limit ${t \rightarrow
  \infty}$ converges to the Poisson solution.  Rewriting terms of
equation \ref{eqn:policy}, this is analogous to a discrete Heat Equation
with source and $\alpha = 1$:
\begin{equation}
V_{k+1} - V_k = r - (I - \gamma P)V_k.
\end{equation}
where the left hand side corresponds to a discrete time derivative,
and the right-hand side corresponds to the Poisson resolvent \cite{TsitsiklisVanRoy}.

In continuous time, letting $L := P - I$, one obtains
\begin{equation}
\frac{d}{dt} u(t) = L u(t) + r - \kappa u(t),
\end{equation}
where $\kappa$ corresponds to discounting. 
Thus, value functions arise as steady states of forced diffusion with decay.

\section{Green's Functions: The Fundamental Matrix and Dayan's Successor Representation}


In a transition matrix, absorbing states $\partial \mathcal{S}$ are
those with no outward transitions, and hence have a single 1 in their
rows.  By reordering states and the corresponding rows and columns in
the transition matrix, a block structure can be constructed for MDPs.
This helps us find a kind of Green's function or equivalently, Dayan's
Successor Representation for RL \cite{Dayan93}.  These
representations allow us to compute value (or Q) given only knowledge
of the reward function.  Although computed slightly differently,
computation of the ``Green's Function'' for episodic and
non-abosorbing MDPs is straightforward.

\subsection{Absorbing (Episodic) MDPs}

Partition states into transient $\mathcal{T}$ and terminal $\partial
\mathcal{S}$ and sort $P$ to group these states so that transient
states are in the upper left block.
The sorted transition matrix then has the following block form:
\begin{equation}
P =
\begin{bmatrix}
Q & B \\
0 & I
\end{bmatrix}.
\end{equation}

In the absorbing case, we can compute value from reward using this relationship:
\begin{equation}
(I - Q)V_\mathcal{T} = B g.
\end{equation}
where $V_\mathcal{T}$ is the value function over transient states and
$g(s)$ is a function over absorbing states that (when normalized to
$[0,1]$) represents outcome (absorption) probabilities that can be
configured depending on the problem to be solved.

\subsection{MDPs with Distributed Reward: The Poisson Case (Step Rewards)}

For ongoing control problems, we consider a non-absorbing chain.
Under this case, $P = Q$ since there are no absorbing states (i.e.,
$Q$ fills $P$).  The discrete Poisson equation then looks like this:

\begin{equation}
(I - P)V = r
\end{equation}

The fundamental matrix is the result of repeated action of Q for
progressively more iterations.  It can be expressed as a Neumann
series that converges to the fundamental matrix $\mathcal{N}$:
\begin{equation}
\mathcal{N} = (I - Q)^{-1} = \sum_{t\ge0} Q^t
\end{equation}
and acts as a Green's function for the Markov chain potential:
\begin{equation}
  V = \mathcal{N} r
\end{equation}

Without other constraints, solutions to Poisson's equation are only
unique up to an additive constant, so fixed-point iteration for this
case can only converge if something like a zero-mean constraint is
imposed.  Define average reward $\bar r$ and differential value $h$:
\begin{equation}
(I - P)h = r - \bar r \mathbf{1}.
\end{equation}
A normalization condition (e.g.\ $h(s_0)=0$) fixes the additive
constant \cite{Puterman94} and enforces uniqueness of the solution
without risking divergence of the solution.

\section{Successor Representation}

Dayan's\cite{Dayan93} successor representation (SR) was developed in
the context of reinforcement learning and resembles the fundamental
matrix with the twist that exponential discounting can be applied to
the transition operator $P$ as follows:

\begin{equation}
M = \sum_{t\ge0} (\gamma P)^t = (I - \gamma P)^{-1}.
\end{equation}
Then
\begin{equation}
V = M r.
\end{equation}

When discounting is applied through the factor $\gamma \in [0,1]$:
\begin{eqnarray}
  (I - \gamma P)V = r  \\
  (I - \gamma P)^{-1} r = V
\end{eqnarray}
This equation has a unique solution for $\gamma<1$ by virtue of
exponential discounting.

\section{Eligibility Trace to Eligibility Field}

Traditional TD distributes credit (the value
update) along recent state traces (trajectories explored during RL
learning).  This is similar to a Monte Carlo solution to the
potential-theory counterpart.  Eligibility traces reveal state-space
topology along chains of experience--the {\em eligibility traces}.
However, the structure of the underlying transition operator can be
learned (estimated) as an RL agent explores state space.  The
potential-theoretic view suggests that with a learned transition
structure, an eligibility {\em field} can be defined that exploits
this learned structure to update a significantly larger set of states
than TD alone can update.

Eligibility distributed over a trace is defined by TD($\lambda$) as:
\begin{equation}
e_t(s) = \gamma \lambda e_{t-1}(s) + \mathbf{1}\{S_t=s\}.
\end{equation}

where $\mathbf{1}\{S_t=s\}$ represents the one-hot encoding of state.

Eligibility may be generalized to a field $e'$ using the partially learned
transition operator $P^\top$:
\begin{equation}
e'_{t+1} = \gamma \lambda P^\top e'_t + \phi(S_{t+1}),
\end{equation}
where $\phi(S_{t+1})$ is an injection vector (e.g., one-hot at 
$S_{t+1}$).  Compared to the classic accumulating trace
$e_{t+1} = \gamma \lambda e_t + \phi(S_{t+1})$,
this inserts a spatial propagation step via 
$P^\top$.
Eligibility not only decays along chains, it can be spread backward
along the {\em graph} defined by $P^\top$ so credit diffuses backward across
the learned Markov transition structure.

\section{Summary and Observations}

This discussion has been limited to RL with a fixed policy.  More
generally, the max operation in policy iteration (and the
max-min operation in game theory) results in nonlinearities that lead
to nonlinear potential theory, which is out of the scope of this
paper.  Nonetheless, reinforcement learning with fixed policies can be
viewed with respect to discrete potential theory:

\begin{itemize}
\item Laplace equation: terminal-only rewards $\Rightarrow$ harmonic functions / hitting probabilities.
\item Poisson equation: rewards as sources.
\item Heat equation: dynamic relaxation toward equilibrium (to Laplace or Poisson).
\item Successor representation: Green's function of the Markov operator.
\end{itemize}
When the state space is large enough, harmonic value functions can be
prone to barren plateaus, especially near saddle points.  Laplace's
equation in particular allows us to imagine optimal state sequences as
{\em discrete streamlines} that maximize goal-reaching success.  A
similar view motivated the use of harmonic functions as potentials for
robot motion planning \cite{Tarassenko91,Connolly92d,Stan94}.

{\bf Formal guarantees for RL:} Episodic RL problems yield value
functions that are harmonic over transient states.  The mean-value and
min-max properties of harmonic functions guarantee that when fully
converged, harmonic value functions should exhibit no local minima.
This also suggests tests for RL convergence and correctness.  The
mean-value property additionally suggests interpolation strategies.
Harmonic interpolants (e.g., harmonic polynomials \cite{Axler91}) can be
employed to interpolate value functions and collapse barren plateaus
in state space.

{\bf Sample Efficiency:} The potential-theoretic viewpoint suggests
possible acceleration of RL: The TD eligibility {\em trace} over a
recent one-dimensional {\em chain} of states becomes a
multidimensional eligibility {\em field} over learned transition
structures, suggesting faster credit distribution over a larger set of
states and thus better sample efficiency.  The viewpoint also admits
analog computing substrates for solving RL problems \cite{Stan94}.
For example, a resistive network effectively computes value functions.
For the episodic (absorbing) case, the TD update rule can be used to
learn (modify) voltages at absorbing states that correspond to
terminal reward probabilities.  The network itself distributes credit
through the learned transition operator.

{\bf Neurobiology:} Although beyond the scope for this paper,
there are intriguing parallels between RL theory as described here and
biological reinforcement learning as observed in midbrain dopamine
signaling.  A major recipient of dopamine signals is a subcortical
structure: the striatum.  This structure also receives immense
convergent projections from the cortex and has long been associated
with action selection.  Dopamine signals in the ventral striatum are
associated with the RPE, which resembles temporal difference in RL
(received minus expected reward \cite{SchultzDM97}.  The RPE as
observed by Schultz and colleagues is a transient pulse of dopamine
delivered to the striatum during trial-and-error or implicit learning
and corresponds closely to the TD training signal at terminal states
of an MDP.  This transient diminishes and ultimately
disappears\footnote{More precisely, the transient migrates backward in time to the
appearance of predictive cues.} during habit formation as the expected
reward approaches actual reward.

The relationship between RL and potential theory raises the possibility
that an electrical network within the striatum might contribute to
fast computation of value functions.  One possible neural substrate
for this computation is the electrically-coupled astrocytic syncytium
surrounding striatal projection neurons \cite{Pai2026}.  Once thought of solely as
glial support for extracellular ionic stability, astrocytes are now
felt to be more directly involved in neural computation
\cite{Santello2019,Abiero2026}.  In the striatum, astrocytes are in a
position to modulate membrane potentials in a way that resembles
state values or Q functions in RL.

\bibliography{mdp}



\end{document}